\pdfoutput=1

\documentclass[preprint,12pt]{elsarticle}

\usepackage{amssymb}
\usepackage{amsmath}
\usepackage{multirow}
\usepackage{makecell}
\usepackage{graphicx}
\usepackage{subfigure}
\usepackage{booktabs}
\usepackage{algorithm}
\usepackage{algpseudocode}
\usepackage{epstopdf}
\usepackage{url}
\biboptions{sort&compress}
\usepackage[nodots,nocompress]{numcompress}

\journal{arXiv}

\begin{document}

\begin{frontmatter}

\title{SocialGuard: An Adversarial Example Based Privacy-Preserving Technique for Social Images}

\author[label1]{Mingfu Xue}
\author[label1]{Shichang Sun}
\author[label2]{Zhiyu Wu}
\author[label1]{Can He}
\author[label1]{Jian Wang}
\author[label3]{Weiqiang Liu}
\address[label1]{College of Computer Science and Technology, Nanjing University of Aeronautics and Astronautics, Nanjing, China}
\address[label2]{College of Science, Nanjing University of Aeronautics and Astronautics, Nanjing, China}
\address[label3]{College of Electronic and Information Engineering, Nanjing University of Aeronautics and Astronautics, Nanjing, China}

\begin{abstract}
  The popularity of various social platforms has prompted more people to share their routine photos online.
    However, undesirable privacy leakages occur due to such online photo sharing behaviors.
    Advanced deep neural network (DNN) based object detectors can easily steal users' personal information exposed in shared photos.
    In this paper, we propose a novel adversarial example based privacy-preserving technique for social images against object detectors based privacy stealing.
    Specifically, we develop an \textit{Object Disappearance Algorithm} to craft two kinds of adversarial social images. One can hide all objects in the social images from being detected by an object detector, and the other can make the customized sensitive objects be incorrectly classified by the object detector. The \textit{Object Disappearance Algorithm} constructs perturbation on a clean social image.
    After being injected with the perturbation, the social image can easily fool the object detector, while its visual quality will not be degraded.
    We use two metrics, privacy-preserving success rate and privacy leakage rate, to evaluate the effectiveness of the proposed method.
    Experimental results show that, the proposed method can effectively protect the privacy of social images. The privacy-preserving success rates of the proposed method on MS-COCO and PASCAL VOC 2007 datasets are high up to 96.1\% and 99.3\%, respectively, and the privacy leakage rates on these two datasets are as low as 0.57\% and 0.07\%, respectively.
    In addition, compared with existing image processing methods (low brightness, noise, blur, mosaic and JPEG compression), the proposed method can achieve much better performance in privacy protection and image visual quality maintenance.

\end{abstract}

\begin{keyword}
Artificial intelligence security, Privacy protection, Social photos, Adversarial examples, Object detectors

\end{keyword}

\end{frontmatter}

\section{Introduction}
In recent years, artificial intelligence (AI) techniques have been widely applied in a variety of tasks.
Although bringing many conveniences, the abuse of AI techniques will also cause a series of security problems, such as social privacy disclosure.
With the popularity of social platforms, people are keen to share their photos online.
However, malicious hackers or entities can leverage the deep neural network (DNN) based object detectors (or image classifiers) to extract valuable private information from those uploaded social photos.
In this way, they can analyze and explore users' preferences, and push the accurate and targeted commercial advertisements \cite{hardt2012privacy}.
Moreover, the leakage of sensitive information (such as identity) can even cause the users to suffer from property losses.
For example, Echizen \cite{online2017japan} indicated that the fingerprint might be stolen from the shared photos that contain the ``V'' gesture.

So far, there are several defense strategies to protect the privacy of the shared social photos.
The most direct and common countermeasure is to set user access control \cite{klemperer2012tag, vishwamitra2017towards, li2019hideme}, where the strangers are not permitted to browse private personal photos.
In addition, the image encryption based techniques \cite{sun2016processing, abdulla2019hitc} are also effective for protecting sensitive data on shared photos.
However, the above two methods will either reduce the users' social experience (access control), or will require huge computation overhead (image encryption).
Therefore, the image processing methods, such as blur, noise and occlusion, have been widely used to protect the privacy contents in images.
However, these image processing methods will greatly affect the visual qualities of the photos.
Moreover, they are ineffective to resist the detection of those advanced DNN based object detectors \cite{liu2017protecting, wilber2016can}.

In this work, we propose a novel adversarial example based method to protect social privacy from being stolen. The key idea is that, the DNN based object detectors are sensitive to small perturbations \cite{liu2017protecting,papernot2018sok,akhtar2018threat}, \textit{i.e.}, adversarial examples.
By constructing adversarial version of these uploaded social photos, \textit{i.e.}, adding some visual invisible perturbations into the photos, the DNN based object detectors can be successfully fooled.
Consequently, the object detectors will make incorrect predictions, or even completely unaware of the existence of the objects in a photo, thus the private information in the social photos can be protected.

The main contributions of this paper are as follows:
\begin{itemize}
  \item
  \textbf{Protect the social privacy effectively without affecting the visual qualities of the photos.}
  We propose a method based on adversarial examples to protect the privacy in social photos, which can make Faster R-CNN \cite{ren2015faster} detector fails to predict any bounding box in a social image.
  Experimental evaluations on two datasets (MS-COCO \cite{lin2014microsoft} and PASCAL VOC 2007 \cite{everingham2010pascal}) show that, the privacy-preserving success rate of the proposed method is high up to 96.1\% and 99.3\%, respectively.
  Compared with existing social privacy protection methods, the proposed method can effectively protect image contents by fooling DNN based object detectors, without affecting the visual qualities of the social photos.
  \item
  \textbf{Supporting customized privacy setting.}
  Some objects (such as person or commercialized objects) in social images contain personal information and thus can be considered as sensitive objects.
  The proposed method can not only hide all the objects from being detected, but can also make the customized sensitive objects in a social image be incorrectly predicted.
  The user can customize the sensitive objects that he wants to keep private.
  After the social image is processed by the proposed method, Faster R-CNN will identify the customized sensitive objects incorrectly.
  \item \textbf{Use the attack method as a defense.}
  Adversarial example used to be a malicious attack method against DNN models.
  In this paper, we are doing the opposite by using the attack as a defense technique for resisting the detection of DNN based object detectors, so as to protect the privacy information in shared social photos from being disclosed.
\end{itemize}

The rest of this paper is organized as follows. Section \ref{Related_work} describes the related work on adversarial example attacks, the workflow of Faster R-CNN, and the existing social privacy protection works against DNN based privacy stealing.
Section \ref{Proposed_method} elaborates the proposed method.
Section \ref{Experiments} presents the experimental results on two standard image datasets.
This paper is concluded in Section \ref{Conclusion}.

\section{Related work} \label{Related_work}
In this section, the adversarial example attacks, the workflow of Faster R-CNN, and the existing four social privacy protection works against DNN based privacy stealing, are reviewed.

\subsection{Adversarial example attacks}
Some researches have indicated that DNN models are vulnerable to adversarial examples, and a range of adversarial example generation methods have been proposed.
Szegedy \textit{et al.} \cite{szegedy2013intriguing} first proposed adversarial example attacks against the DNN models, and developed the \textit{L-BFGS algorithm} to construct the adversarial perturbations.
By adding the generated perturbations into a clean image, the target DNN model will output the incorrect classification result.
Then, Goodfellow \textit{et al.} \cite{goodfellow2014explaining} proposed the \textit{Fast Gradient Sign Method} (FGSM) to explain the linear behavior of adversarial perturbations, which is effective to accelerate the generation of adversarial examples.
Carlini and Wagner \cite{carlini2017towards} constructed the high-confidence adversarial examples with three different distance metrics (\textit{i.e.}, ${L_0}$ distance, ${L_2}$ distance, and ${L_\infty}$ distance), and their generated adversarial examples could successfully fool the distilled neural networks.
Chen \textit{et al.} \cite{chen2018shapeshifter} developed the \textit{ShapeShifter} attack to craft physical adversarial examples to fool road sign object detectors.
By applying the \textit{Expectation over Transformation} technique \cite{abs-1712-09665,athalye2017synthesizing}, the \textit{ShapeShifter} is robust to the changes of different distances and angles.

\subsection{Workflow of Faster R-CNN}
The workflow of the Faster R-CNN detector includes two stages \cite{ren2015faster}: region proposal and box classification.
In the first stage, the features in an image are extracted, and the region proposal network (RPN) is used to determine where an object may exist and mark the region proposal.
In the second stage, the corresponding objects in the marked region proposals are classified by DNN based classifier.
Meanwhile, the \textit{bounding box regression} \cite{girshick2015fast} is performed to obtain the accurate location of each object in the image.

\subsection{Social privacy protection against DNN based privacy stealing}
We proposed the idea of this paper in 2019 and applied for a Chinese patent \cite{ourpatent}.
Besides, few exploratory studies have been conducted on the protection of image privacy against DNN based privacy stealing.
Li \textit{et al.} \cite{li2019scene} presented the \textit{Private Fast Gradient Sign Method} (P-FGSM) to prevent the scene (e.g., hospital, church) in social images from being detected by DNN model.
They added imperceptible perturbation to the social images to generate the adversarial social images, in which the scene will be incorrectly classified by DNN classifier.
Shen \textit{et al.} \cite{ShenFWNK19} proposed an algorithm to protect the sensitive information in social images against DNN based privacy stealing and keep the changes of the social images imperceptible.
They also conducted a human subjective evaluation experiment to evaluate the factors that influence the visual sensitivity of human \cite{ShenFWNK19}.
The difference between our work and \cite{ShenFWNK19} is that, \cite{ShenFWNK19} focuses on preventing sensitive information from being detected by DNN based classifier, while our work focuses on preventing sensitive objects from being detected by object detector.
Shan \textit{et al.} \cite{abs-2002-08327} proposed a \textit{Fawkes} system to protect personal privacy against unauthorized face recognition models.
Specifically, they add pixel-level perturbations to user's photos before uploading them to the Internet \cite{abs-2002-08327}.
The functionality of unauthorized facial recognition models trained on these photos with perturbations will be deteriorated seriously.
The differences between our work and \cite{abs-2002-08327} are that: (i) The work \cite{abs-2002-08327} protects social privacy against facial recognition models, while our method protect social privacy against object detectors. (ii) The work \cite{abs-2002-08327} adds adversarial images to the training set of unauthorized facial recognition models to influence the training phase of these models (which is a strong assumption), while our work focuses on the testing phase of DNN detectors.

The above three related works focus on privacy protection against the image classifiers.
Compared with image classifiers, attacking or deceiving object detectors is more challenging \cite{zhao2018seeing,eykholt2018physical,liu2017protecting}.
The reasons are as follows.
First, an object detector is more complex, which can detect multiple objects at once.
Second, the object detector can infer the true objects by the obtained information from the image background.
To the best of the authors' knowledge, the only social privacy protection work against object detectors is Liu \textit{et al.} \cite{liu2017protecting}.
Liu \textit{et al.} \cite{liu2017protecting} constructed the adversarial examples with a stealth algorithm, which made all objects in the image invisible to the object detectors, thus protecting the image privacy from being detected.
Compared to the work \cite{liu2017protecting}, our work has the following differences:
(i) The mechanisms are different. The work \cite{liu2017protecting} focuses on the region proposal stage (the first stage) of Faster R-CNN, while our method focuses on the classification stage (the second stage) of Faster R-CNN.
Moreover, our method leverages the classification boundary of object detector to craft perturbations.
Specifically, the position of the feature point of each bounding box is moved and crossed the classification boundary, which make object detector cannot find any sensitive object and abandon the corresponding bounding boxes, while the stealth algorithm in \cite{liu2017protecting} performs a backpropagation on the loss function to create perturbations and suppress the generation of bounding boxes.
(ii) The effects are different. The work \cite{liu2017protecting} can hide all the objects in a social image from being detected, while our method can not only achieve this, but can also make the customized sensitive objects in a social image be misclassified as other classes.

\section{Proposed adversarial example based privacy-preserving technique} \label{Proposed_method}
In this section, the proposed adversarial example based social privacy-preserving technique is elaborated.
First, the privacy protection goals for social photos are described.
Second, the overall flow of the proposed method is presented.
Third, the \textit{Object Disappearance Algorithm} is elaborated.
The proposed method can generate two kinds of adversarial examples.
One can make all the objects in a social image invisible, and the other can make the customized sensitive objects in a social image be incorrectly predicted.

\subsection{Privacy goal} \label{Pivacy goal}
DNN based object detectors perform excellently in various computer vision tasks.
The deep learning based object detectors can be divided into two categories:
(i) one-stage object detectors, such as You Only Look Once (YOLO) \cite{redmon2016you}, and Single Shot Multibox Detector (SSD) \cite{liu2016ssd};
(ii) two-stage object detectors, such as Faster R-CNN \cite{ren2015faster}, and Region-based Fully Convolutional Networks (R-FCN) \cite{dai2016r}.
Existing research \cite{huang2017speed} has indicated that, compared to SSD and R-FCN, the Faster R-CNN can achieve more superior performance on the trade-off between speed and detection accuracy. Besides, generally, Faster R-CNN has higher detection accuracy than that of YOLO detector.
Therefore, in this paper, the proposed method is targeting at protecting image privacy from being detected by Faster R-CNN detectors.

The privacy goal of the proposed method includes two aspects: one is that Faster R-CNN fails to predict any bounding box in an image, which will be discussed in Section \ref{novel adv}, and the other is that Faster R-CNN can predict some bounding boxes in an image but classifies the sensitive objects incorrectly, which will be discussed in Section \ref{Privacy settings}.
If one of these two aspects is met, it is considered that the social privacy is successfully protected.
To achieve the first aspect, we need to make sure that for each bounding box, the score of each bounding box is lower than the threshold, which can be formalized as:
\begin{equation}
\label{equ1}
\max[P_{k}(x')] < T \;\;(k=1,2,...,K)
\end{equation}
where $x'$ is an adversarial social image. $K$ is the number of classes that an object detector can recognize ($K=80$ in the experiment), and $k$ denotes the \textit{k}-th category. $P_k$ is the score of class $k$ of the bounding box. $T$ is the detection threshold of the system.

For an object detector, it outputs bounding boxes with high scores (higher than the threshold), and discards the bounding boxes with low scores (lower than the threshold).
Equation \eqref{equ1} indicates that the scores predicted by Faster R-CNN for all the bounding boxes are all lower than the threshold. Therefore, the system will abandon all bounding boxes and detect nothing in the adversarial social image.

The proposed adversarial example based method can make all objects in a social image undetectable by Faster R-CNN (the first aspect), which will be discussed in Section \ref{novel adv}.
In addition, the proposed method can also make the customized sensitive objects incorrectly classified as another object (the second aspect), which will be discussed in Section \ref{Privacy settings}.

\subsection{Overall flow}
\begin{figure*}[!htbp]
	\centering
	\includegraphics[width=\textwidth]{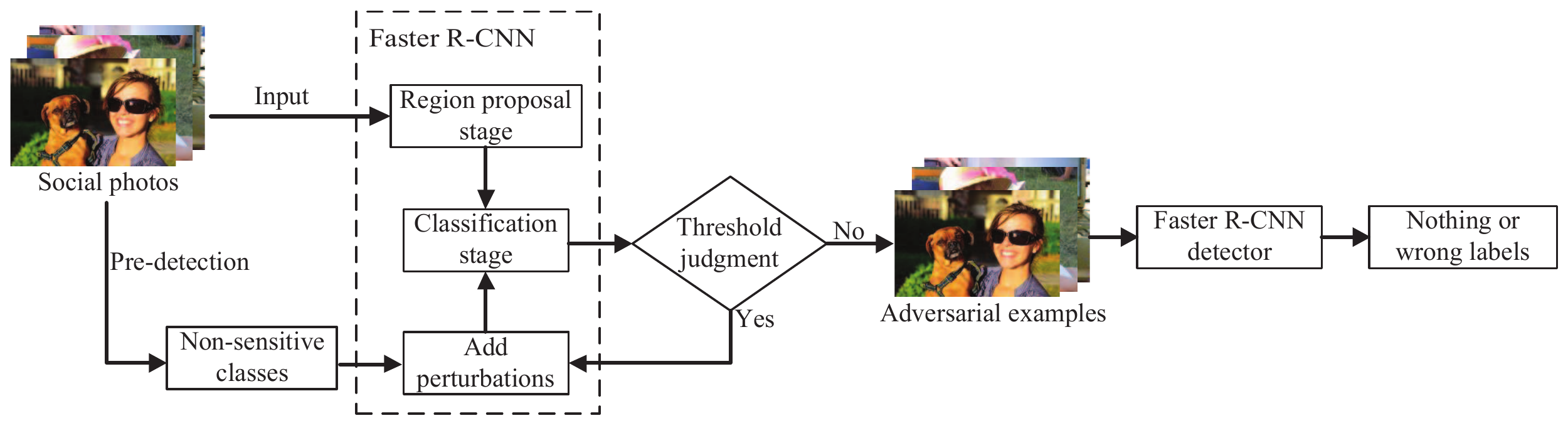}
	\caption{The overall flow of the proposed adversarial example based privacy-preserving technique.}
	\label{fig1}
\end{figure*}

The overall flow of the proposed method is shown in Fig. \ref{fig1}.
The method takes users' photos as the inputs, and outputs the adversarial examples to deceive the Faster R-CNN object detector.
First, pre-detection is performed on the users' photos to obtain the detected objects.
Then, the classes that do not belong to those correctly detected classes are selected as non-sensitive classes (defined in Section \ref{novel adv}).
Second, the region proposals ${r_1},{r_2},...,{r_m}$ are obtained in the classification stage of Faster R-CNN.
Third, the adversarial attack is implemented at the second stage (\textit{i.e.}, the classification stage) of Faster R-CNN by iteratively constructing the perturbation in the image.
The perturbation is iteratively constructed until that, for every bounding box, the score of each object is lower than the threshold, which will cause Faster R-CNN to discard all the bounding boxes.
In this way, the generated adversarial social image can successfully deceive the Faster R-CNN detector when it is uploaded to the social platforms.

\subsection{Object Disappearance Algorithm}\label{novel adv}
We propose an \textit{Object Disappearance Algorithm} which is presented in Algorithm \ref{alg1}, to generate adversarial examples and prevent Faster R-CNN from predicting any bounding box (the first aspect as discussed in Section \ref{Pivacy goal}).
The proposed method focuses on attacking the classification network at the second stage (\textit{i.e.}, the classification stage) of Faster R-CNN.
The adversarial attack includes two steps.
First, the pre-detection on the original image $x$ is performed to obtain the predicted classes, including the correct and incorrect prediction results.
We define the correctly predicted classes as the sensitive classes that need to be protected.
In this paper, for an unprocessed social image, the sensitive objects are the objects that appear in the image and can be detected by object detector, and the non-sensitive objects are the objects that do not appear in the image.
For different social images, the sensitive objects and the non-sensitive objects are different.
Then, the non-sensitive class set ${Y'} = \{ {y'_1},{y'_2},...,{y'_{K-d}}\}$ is selected, where $K$ is the number of classes that an object detector can recognize, and $d$ is the number of classes that are detected in the original image.
${y'_1},{y'_2},...,{y'_{K-d}}$ do not belong to these correctly detected classes and are the classes of objects that do not appear in the original image $x$.
Second, the region proposals ${r_1},{r_2},...,{r_m}$ are obtained in the classification stage of Faster R-CNN, and the perturbations are constructed iteratively in the original image $x$ with the following formula:
\begin{equation}
\label{equ2}
\begin{aligned}
\mathop {\arg \min }\limits_{x{'_i}} {\kern 1pt} &\frac{1}{m}\sum\limits_{{r_j}} {L(F{{(x{'_i})}},y{'_{t}})}
\end{aligned}
\end{equation}
where $x{'_i}$ is an adversarial example generated in the $i$-th iteration.
$r_j$ represents the {\textit{j}-th predicted bounding box, and the total number of bounding boxes is $m$.
${y'_{t}}$ is the label of the non-sensitive class and ${y'_{t}} \in {Y'}$.
$y'_{t}$ is not a fixed class label and is sequentially selected from the set ${Y'}$ in each iteration.
$F$ represents the classification network of Faster R-CNN, and $L$ is the loss function that computes the difference between the output $F(x{'_i})$ of the classification network and the non-sensitive class label $y'_{t}$.

\renewcommand{\algorithmicrequire}{\textbf{Input:}}
\renewcommand{\algorithmicensure}{\textbf{Output:}}
\begin{algorithm}[!htb]
  \caption{Object Disappearance Algorithm.}
  \label{alg1}
  \begin{algorithmic}[1]
    \Require
      original image $x$; perturbation constraint $\varepsilon$; detection threshold $T$; maximum number of iterations $I$;
    \Ensure
       adversarial example $x'$;
    \State \textbf{Initialize} $x{'_0} = x$; $i = 1$;
    \State Conduct a pre-detection on $x$ to obtain the classified objects;
    \State Select the non-sensitive class set $Y{'}$;
    \State Obtain the region proposals ${r_1},{r_2},...,{r_m}$ in the classification stage of Faster R-CNN.
    \While {$i < I$}
      \State Add perturbation to the original image $x$ and get the adversarial
      \Statex \quad \; example $x{'_i}$;
      \State Clamp($x{'_i}$, $x{'_{i-1}}-\varepsilon$, $x{'_{i-1}}+\varepsilon$);
      \State Use the classification network to predict the class of objects for all
      \Statex \quad \; the bounding boxes in $x{'_i}$;
      \State Get the scores $S$ of each class for all the bounding boxes;
      \State ${s_{max}} = \max ({S})$;
      \If {${s_{max}} < T$}
        \State $x' = x{'_i}$;
        \State break;
      \EndIf
      \State $i = i + 1$;
    \EndWhile \\
    \Return $x'$;
  \end{algorithmic}
\end{algorithm}

In each iteration, we sequentially select the non-sensitive class ${y'_{t}}$ from the set ${Y'}$.
Then, we construct the adversarial perturbation with formula \eqref{equ2} and utilize the constraint $||x{'_i} - x{'_{i-1}}|| < \varepsilon$ to make the perturbation visually invisible, where $\varepsilon$ restricts the intensity of the generated perturbations.
After the adversarial example $x{'_i}$ for the $i$-th iteration is generated, the classification network $F$ is used to predict the classes of objects for all the bounding boxes ${r_1},{r_2},...,{r_m}$.
Subsequently, the scores ${S}$ of each class for all the bounding boxes ${r_1},{r_2},...,{r_m}$ will be obtained. $S$ can be considered as a matrix as follows:
\begin{equation}
\label{equ3}
S = \left( {\begin{array}{*{20}{c}}
{{s_{11}}}&{{s_{12}}}& \cdots &{{s_{1K}}}\\
{{s_{21}}}&{{s_{22}}}& \cdots &{{s_{2K}}}\\
 \vdots & \vdots & \ddots & \vdots \\
{{s_{m1}}}&{{s_{m2}}}& \cdots &{{s_{mK}}}
\end{array}} \right)
\end{equation}
where $s_{ij}$ is the score of class $j$ for the $i$-th bounding box ($i=1,2,...,m$, $j=1,2,...,K$).
The maximum score ${s_{max}}$ is then calculated, where ${s_{max}}$ represents the maximum score of all classes for all the bounding boxes ${r_1},{r_2},...,{r_m}$ in the generated adversarial example $x{'_i}$.
We denote the maximum number of iterations as $I$.
The iteration process is terminated when the iteration number reaches $I$ or ${s_{max}}$ is lower than the threshold, which means that there is no bounding box that can be detected by Faster R-CNN.

Here, we explain the reason why the above algorithm can cause all objects in the generated adversarial example invisible.
For sensitive classes in an image, adding perturbation with Equation \eqref{equ2} can change the classification boundary of sensitive classes and reduce the scores of them. Thus, the scores of sensitive objects are below the detection threshold.
For non-sensitive classes, the multi-classes perturbing method restraints the increase of non-sensitive objects' scores with insufficient iterations. As a result, the scores of non-sensitive objects are also below the detection threshold.
As a result, the score of each class for any bounding boxes will be lower than the threshold, which makes Faster R-CNN discard all the bounding boxes and detect nothing in the adversarial social image.

\subsection{Customized privacy setting} \label{Privacy settings}

In this section, we describe another method (denoted as $M_{sen}$) of generating adversarial examples which can protect the customized sensitive objects in the social images.
In a social photo, some objects are usually sensitive and private, such as: (i) Person. The person in a social photo may involve users' personal information. For example, a common V-shaped gesture photo may contain the fingerprint information of an user.
(ii) Commercialized objects. Some commercial objects (such as handbags, sports balls, and books) may reveal users' interests. The leakage of these commercialized objects may bring unnecessary advertising to users.
To avoid the disclosure of the sensitive objects in the shared social photos, the proposed method can also provide users with privacy settings, where users can customize the sensitive objects that he wants to protect and the specific class $y_{non}$ that he wants the sensitive objects to be misclassified as.
The proposed method can achieve this purpose by iteratively constructing perturbations on social images with the following formula:
\begin{equation}
\label{equ4}
\begin{aligned}
\mathop {\arg \min }\limits_{x{'_i}} {\kern 1pt} &\frac{1}{m}\sum\limits_{{r_j}} {L(F{{(x{'_i})}},y{_{non}})}
\end{aligned}
\end{equation}

The perturbation is iteratively constructed until that, for every bounding box, the scores of the customized sensitive objects are lower than the threshold.
$M_{sen}$ can cause the score of the class $y_{non}$ to be higher than the threshold, which will make Faster R-CNN classify the customized sensitive objects as the class $y_{non}$.
As a result, Faster R-CNN cannot detect the customized sensitive object and will present the bounding boxes with wrong class (the non-sensitive class $y_{non}$).

We denote the method introduced in Section \ref{novel adv} as $M_{all}$, and the method in this section as $M_{sen}$.
We use $AE_{all}$ and $AE_{sen}$ to represent the adversarial examples generated by $M_{all}$ and $M_{sen}$, respectively.
The difference between the workflows of $M_{all}$ and $M_{sen}$ is shown in Fig. \ref{fig2}. $M_{all}$ performs pre-detection on the original social image to obtain the non-sensitive classes and utilizes multiple non-sensitive classes to generate adversarial example, while $M_{sen}$ does not perform pre-detection and utilizes a fixed non-sensitive class $y_{non}$ to generate adversarial example.

\begin{figure*}[!htbp]
	\centering
	\includegraphics[width=5.5in]{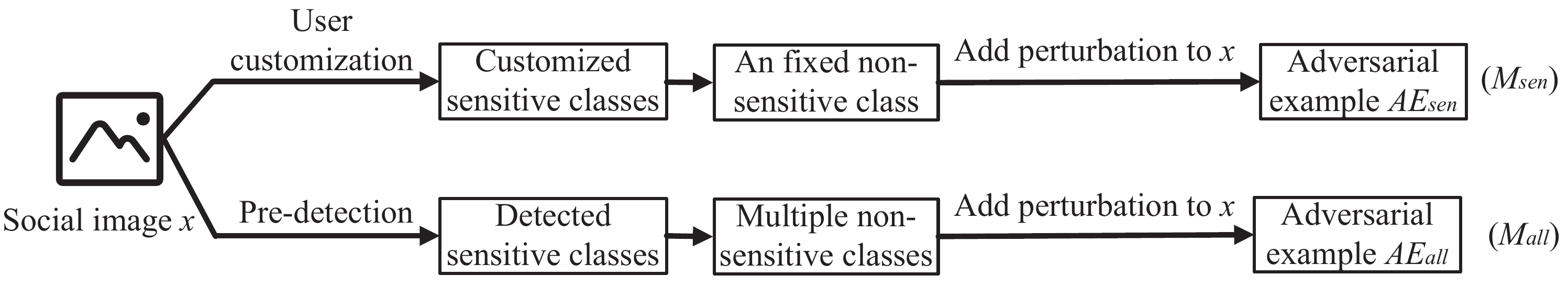}%
	\caption{The difference between the workflows of $M_{all}$ and $M_{sen}$.}
	\label{fig2}
\end{figure*}

In a word, this paper provides two adversarial example based method ($M_{all}$ and $M_{sen}$) to protect the privacy of social photos.
$M_{all}$ hides all objects in a social image from being detected by Faster R-CNN, while $M_{sen}$ is used to protect the sensitive objects customized by the users.

\section{Experimental results} \label{Experiments}
In this section, we evaluate the proposed method.
First, the experimental settings, the experimental datasets, and the two evaluation metrics, are described in Section \ref{Ex_setup}.
Second, we demonstrate the effectiveness of the proposed method in terms of two metrics: the privacy-preserving success rate (Section \ref{Ex_succ}) and the privacy leakage rate (Section \ref{plr}).
Then, we explore the impacts of two parameters (perturbation constraint $\varepsilon$ and detection threshold $T$) on the proposed method in Section \ref{Ex_para}.
Finally, we compare the proposed method with the existing image processing methods in Section \ref{Ex_com}.

\subsection{Experimental setup} \label{Ex_setup}
In our experiments, we evaluate the performance of the proposed method on the Faster R-CNN \cite{ren2015faster} Inception v2 \cite{szegedy2016rethinking} model.
The model has been pre-trained on the Microsoft Common Objects in Context (MS-COCO) dataset \cite{lin2014microsoft} with Tensorflow \cite{abadi2016tensorflow} platform.
We conduct the experiment on MS-COCO dataset \cite{lin2014microsoft} and PASCAL VOC 2007 dataset \cite{everingham2010pascal}.
The MS-COCO dataset has more than 200,000 images for object detection.
It contains 80 object categories including people, vehicles, animals, sports goods, and other common items \cite{lin2014microsoft}.
The PASCAL VOC 2007 dataset \cite{everingham2010pascal} has 20 categories, most of which are the same as the categories in MS-COCO dataset.
We randomly select 1,000 images from MS-COCO dataset \cite{lin2014microsoft}, and 1,000 images from PASCAL VOC 2007 dataset \cite{everingham2010pascal}, respectively, to evaluate the proposed method.

We use two metrics, named privacy-preserving success rate and privacy leakage rate, to evaluate the proposed method on protecting the privacy of social images.
As mentioned in Section \ref{novel adv} and Section \ref{Privacy settings}, the proposed social privacy-preserving method can generate two kinds of adversarial examples.
As discussed in Section \ref{Privacy settings}, we use $AE_{all}$ to represent the adversarial examples that aim to make all objects invisible, and $AE_{sen}$ to represent the adversarial examples that aim to hide the customized sensitive object.
The privacy-preserving success rate ($R_{all}$) for adversarial examples $AE_{all}$ can be formalized as:
\begin{equation}
{R_{all}} = \frac{N_{all}}{N}
\end{equation}
where $N_{all}$ is the number of adversarial social images $AE_{all}$ in which Faster R-CNN cannot present any bounding boxes. $N$ is the number of all the adversarial social images.
The privacy-preserving success rate ($R_{sen}$) for adversarial examples $AE_{sen}$ can be formalized as:
\begin{equation}
{R_{sen}} = \frac{N_{sen}}{N}
\end{equation}
where $N_{sen}$ is the number of adversarial examples $AE_{sen}$ in which Faster R-CNN classifies the customized sensitive class incorrectly.

To further explore the effectiveness of the proposed method on preventing the privacy leakage, we use a metric named privacy leakage rate. The privacy leakage rate represents how many sensitive contents are detected by Faster R-CNN in social images.
The privacy leakage rate $P_{all}$ for the adversarial social images that aim to hide all the objects ($AE_{all}$) can be calculated as follows:
\begin{equation}
\label{equ6}
P_{all} =\frac{A}{O}
\end{equation}
where $A$ is the number of bounding boxes that are detected in all the adversarial social images $AE_{all}$, including the correct and incorrect detection results.
$O$ is the number of bounding boxes that are detected in all the original social images, including the correct and incorrect detection results.
The privacy leakage rate $P_{sen}$ for the adversarial social images that aim to hide the customized sensitive objects ($AE_{sen}$) can be expressed as:
\begin{equation}
P_{sen} =\frac{A_{sen}}{O_{sen}}
\end{equation}
where $A_{sen}$ is the number of bounding boxes of the customized sensitive classes that are correctly detected in the adversarial social images $AE_{sen}$.
$O_{sen}$ is the number of bounding boxes of customized sensitive classes that are correctly detected in the original social images.

\subsection{Privacy-preserving success rate} \label{Ex_succ}
In this section, we use the privacy-preserving success rate to evaluate the proposed method.
Fig. \ref{fig3} shows four examples to illustrate the effectiveness of the proposed method.
The four examples consist of the original images, the detection results of the original images, the detection results of adversarial examples $AE_{all}$, and the detection results of adversarial examples $AE_{sen}$, respectively.
Faster R-CNN can easily recognize all the sensitive objects if the shared images are unprocessed, as shown in Fig. \ref{oriimg} and Fig. \ref{oriimgd}.
Specifically, all the persons are accurately detected in Fig. \ref{oriimgd}.
However, Faster R-CNN detects nothing in the adversarial examples $AE_{all}$, as shown in Fig. \ref{ae1d}.
Meanwhile, it is hard for humans to observe the differences between adversarial examples $AE_{all}$ and the original images.
In Fig. \ref{ae2d}, all the persons in adversarial examples $AE_{sen}$ are misclassified as other objects.
The above results indicate that, the proposed method can protect the social privacy effectively without affecting the visual effect of social images.

\begin{figure*}[!t]
	\centering
	\subfigure[]{\includegraphics[width=0.21\textwidth]{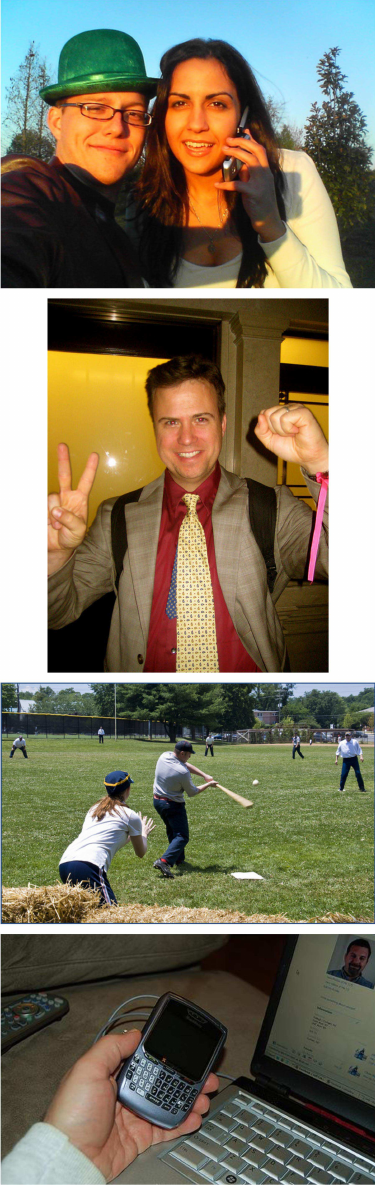}
    \label{oriimg}
	}
	\quad
	\subfigure[]{\includegraphics[width=0.21\textwidth]{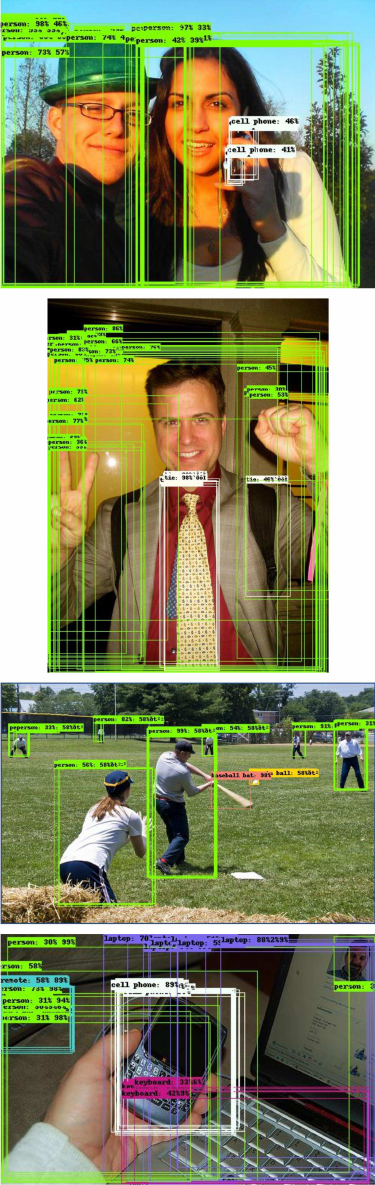}
    \label{oriimgd}
	}
	\quad
	\subfigure[]{\includegraphics[width=0.21\textwidth]{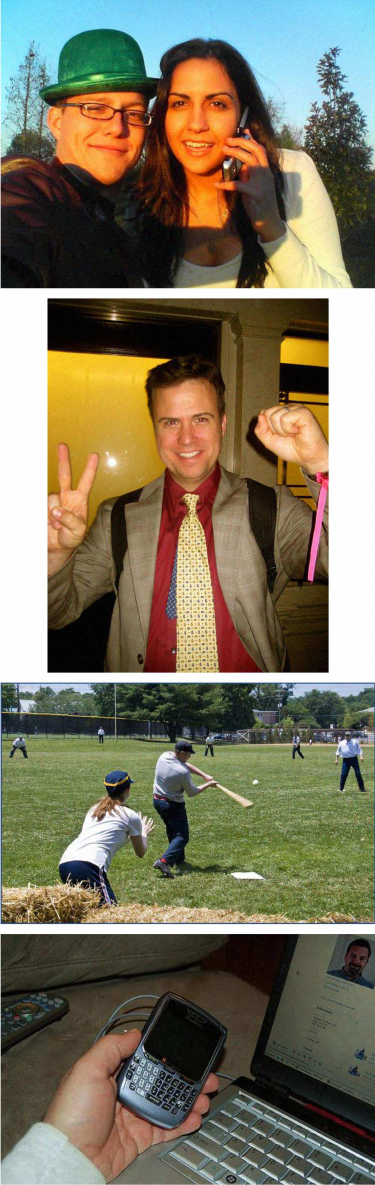}
    \label{ae1d}
	}
	\quad
	\subfigure[]{\includegraphics[width=0.21\textwidth]{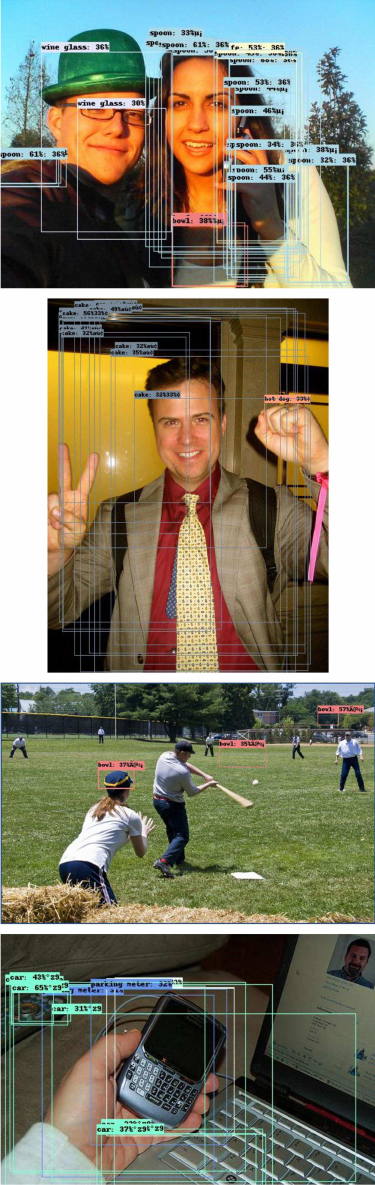}
    \label{ae2d}
	}	
	\caption{Image examples. (a) Original images; (b) the detection results of original images; (c) the detection results of $AE_{all}$; (d) the detection results of $AE_{sen}$.}
	\label{fig3}
\end{figure*}

The privacy-preserving success rates of the proposed method are shown in TABLE \ref{tab1}.
In the experiment, we select 1,000 images in each dataset and test the privacy-preserving success rates for the adversarial social image $AE_{all}$ and $AE_{sen}$, respectively.
The success rates of $AE_{all}$ on MS-COCO and PASCAL VOC 2007 datasets are 96.1\% and 99.3\%, respectively. The success rates of $AE_{sen}$ on the two datasets are 96.5\% and 99.4\%, respectively.
The privacy-preserving success rates of $AE_{all}$ and $AE_{sen}$ are very close and both reach a very high level.
The results indicate that the proposed method performs well on protecting the privacy of social photos.

\begin{table}[!htbp]
  \centering
  \caption{Privacy-preserving success rates of the proposed method on MS-COCO and PASCAL VOC 2007 datasets. $AE_{all}$ represents the adversarial examples that aim to protect all objects. $AE_{sen}$ represents the adversarial examples that aim to protect the customized sensitive object.}
    \begin{tabular}{|c|m{6em}|c|}
    \hline
    \multirow{2}{*}{Dataset}&  \multicolumn{2}{c|}{Privacy-preserving success rate}\cr\cline{2-3}
        & \makecell[c]{$AE_{all}$} & \makecell[c]{$AE_{sen}$}\cr
    \hline
    \hline
    MS-COCO \cite{lin2014microsoft} & \makecell[c]{96.1\%} & 96.5\% \cr\hline
    PASCAL VOC 2007 \cite{everingham2010pascal} & \makecell[c]{99.3\%} & 99.4\%  \cr\hline
    \end{tabular}%
  \label{tab1}%
\end{table}%

\subsection{Privacy leakage rate} \label{plr}
In this section, we use privacy leakage rate to evaluate the proposed method.
We calculate the privacy leakage rates for $AE_{all}$ and $AE_{sen}$ on MS-COCO and PASCAL VOC 2007 datasets, respectively.
As shown in TABLE \ref{tab2}, the privacy leakage rates for adversarial social images $AE_{all}$ on the two datasets are 0.57\% and 0.07\%, respectively, and the privacy leakage rates for adversarial social images $AE_{sen}$ on the two datasets are 2.23\% and 0.18\%, respectively.
The privacy leakage rates of $AE_{all}$ and $AE_{sen}$ both reach a very low level.
The results indicate that, our method can protect the sensitive objects from being detected by Faster R-CNN.

\begin{table}[!htbp]
  \centering
  \caption{Privacy leakage rates of two kinds of adversarial examples on MS-COCO and PASCAL VOC datasets.}
    \begin{tabular}{|c|m{4em}|c|}
    \hline
    \multirow{2}{*}{Dataset} & \multicolumn{2}{c|}{Privacy leakage rate}\cr\cline{2-3}
        & \makecell[c]{$AE_{all}$} & \makecell[c]{$AE_{sen}$} \cr
    \hline
    \hline
    MS-COCO \cite{lin2014microsoft} & \makecell[c]{0.57\%} & 2.23\% \cr\hline
    PASCAL VOC 2007 \cite{everingham2010pascal} & \makecell[c]{0.07\%} & 0.18\% \cr\hline
    \end{tabular}%
  \label{tab2}%
\end{table}%

Considering that our method may have different privacy protection effects for different sensitive objects, we select 9 different sensitive objects and calculate the privacy leakage rates of these objects, respectively.
The 9 classes are person, elephant, airplane, laptop, traffic light, book, sports ball, parking meter, and cell phone.
In the experiment, we select 300 images from MS-COCO dataset \cite{lin2014microsoft} for each class.
Fig. \ref{fig4} shows the privacy leakage rates of the nine sensitive objects in the adversarial social images $AE_{sen}$.
The result indicates that, the proposed method can effectively prevent the privacy information in the social image from being detected. The privacy leakage rates of the nine sensitive classes are all in a very low level (lower than 2\%).
Moreover, the privacy leakage rate of the elephant class is as low as 0.26\%.

\begin{figure}[!htbp]
	\centering
	\includegraphics[width=3.3in]{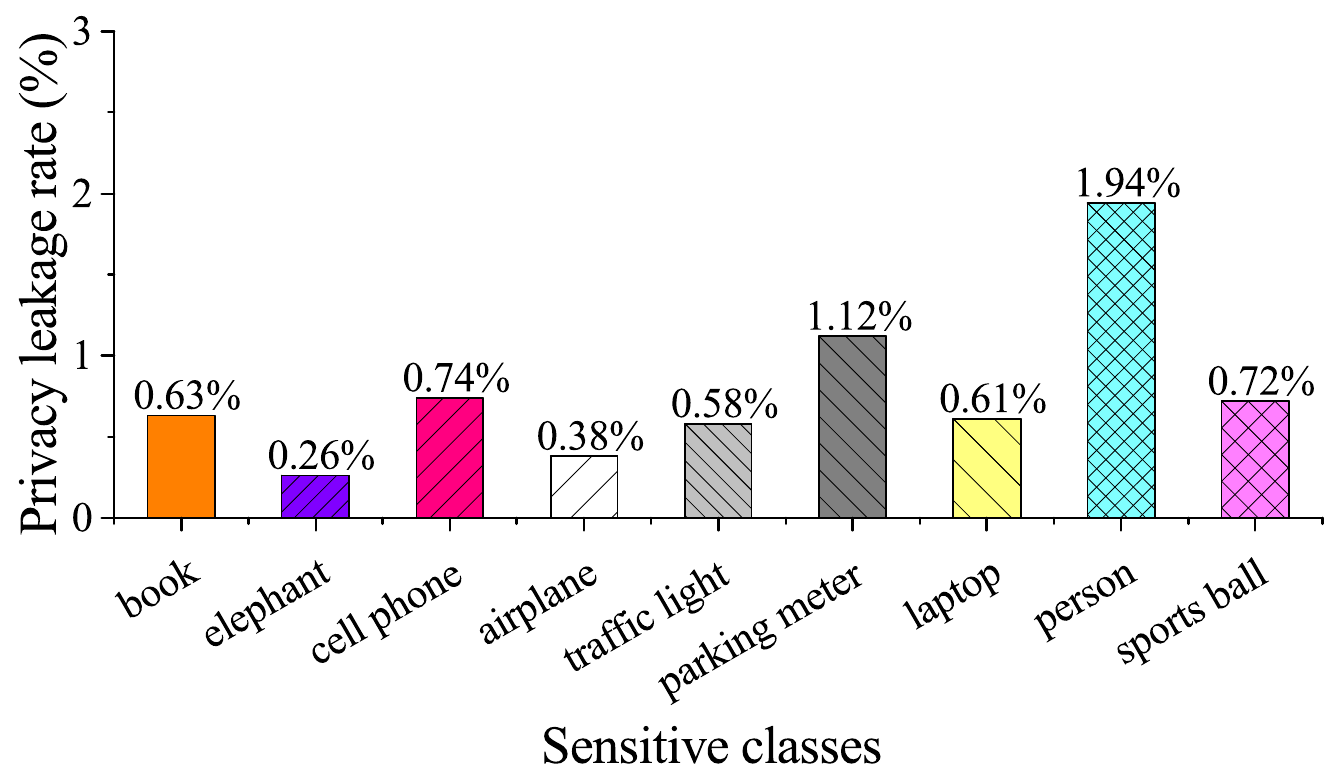}		
	\caption{Privacy leakage rates of $AE_{sen}$ on nine sensitive object classes.}
	\label{fig4}
\end{figure}

\subsection{Parameter discussion} \label{Ex_para}
In the proposed method, there are two parameters that may affect the performance: the perturbation constraint $\varepsilon$ and the detection threshold $T$.
The perturbation constraint $\varepsilon$ constrains the intensity of the perturbation, which makes the difference between the original image and the adversarial example difficult to be noticed. The detection threshold $T$ determines the number of bounding boxes that are presented by Faster R-CNN.

We set the values of the perturbation constraint $\varepsilon$ to be $1/255$$\sim$$10/255$ because these values of $\varepsilon$ are small enough and can make the generated perturbations difficult to be noticed.
The privacy-preserving success rate ${R}_{sen}$ and the privacy leakage rate ${P}_{sen}$ of adversarial social images $AE_{sen}$ under different $\varepsilon$ settings are shown in Fig. \ref{fig5}.
It can be seen that when $\varepsilon  < 3/255$, both the privacy-preserving success rate ${R}_{sen}$ and the privacy leakage rate ${P}_{sen}$ change drastically as $\varepsilon$ increases, and ${R}_{sen}$ and ${P}_{sen}$ tend to stabilize when $\varepsilon  > 3/255$.
Moreover, when $\varepsilon = 3/255$, the value of ${R}_{sen}$ already reaches a high level (96.5\%) and the value of ${P}_{sen}$ is only 2.52\%.
Note that, the larger the value of $\varepsilon$ is, the larger the perturbation strength to an image. A large perturbation strength may affect the visual quality of the perturbed image.
Therefore, $\varepsilon  = 3/255$ has the optimal performance when considering the trade-off between privacy preserving effect and perturbation strength.

We also study the relationship between $\varepsilon$ and the image quality.
Structural Similarity (SSIM) \cite{WangBSS04} and Peak Signal to Noise Ratio (PSNR) are two quantitative metrics to reflect the visual quality of the processed images.
PSNR describes the pixel level difference between two images, while SSIM \cite{WangBSS04} shows structural changes of a processed image.
The higher the value of PSNR and SSIM \cite{WangBSS04}, the more similar between the original image and the modified image.
Fig. \ref{fig6} presents the PSNR and SSIM of the modified images $AE_{sen}$ under different settings of perturbation constraint $\varepsilon$.
Both PSNR and SSIM show downward trends with the increase of $\varepsilon$.
In other words, the smaller the value of $\varepsilon$, the better the visual quality of the image.
Therefore, the value of $\varepsilon$ should be as small as possible to maintain a good visual quality for the adversarial example.
In conclusion, $\varepsilon  = 3/255$ is supposed to be the optimal one when considering the trade-off between image quality and privacy leakage rate.

\begin{figure}[!htbp]
	\centering
    \includegraphics[width = 3in]{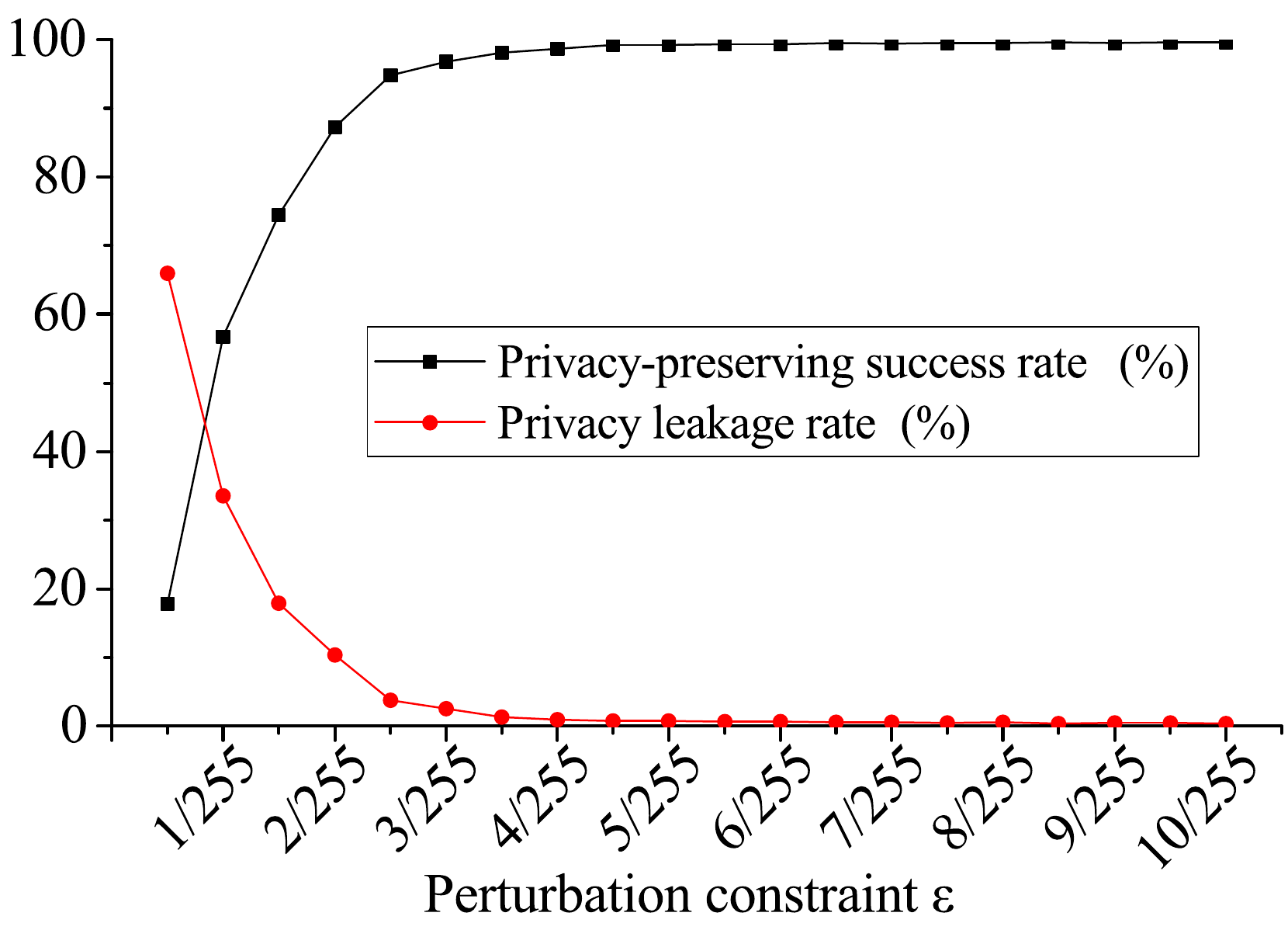}
    \caption{The privacy-preserving success rate and the privacy leakage rate of adversarial social images $AE_{sen}$ under different perturbation constraint $\varepsilon$.}
    \label{fig5}
\end{figure}

\begin{figure}[!htbp]
	\centering
    \includegraphics[width = 3in]{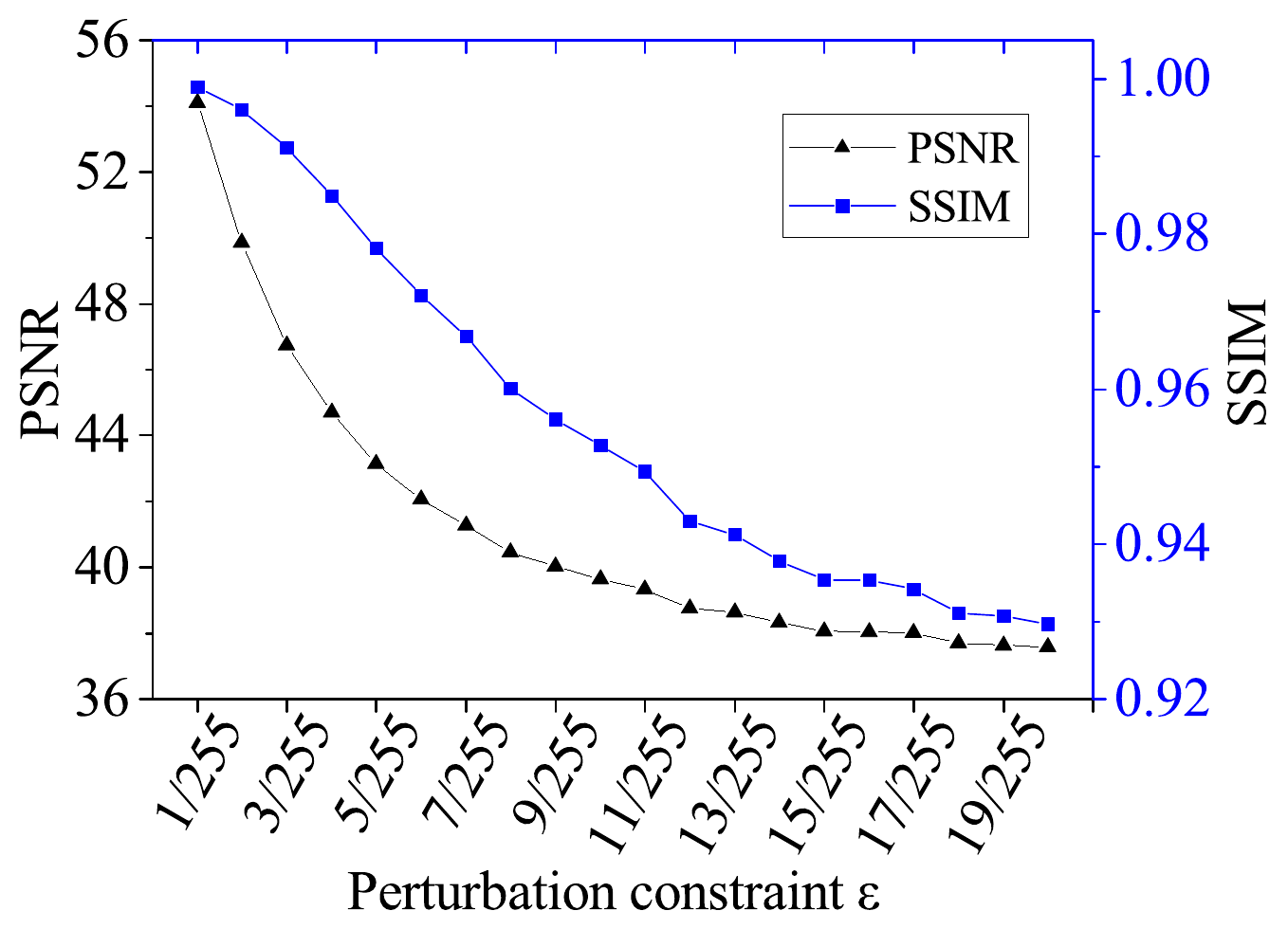}
    \caption{PSNR and SSIM between original images and the adversarial examples $AE_{sen}$ under different perturbation constraint $\varepsilon$.}
	\label{fig6}
\end{figure}

We further explore the impact of the detection threshold $T$ on the privacy-preserving success rate and privacy leakage rate.
The privacy-preserving success rates of $AE_{all}$ and $AE_{sen}$ under different $T$ are shown in Fig. \ref{fig7}.
It can be seen that the success rate of $AE_{all}$ increases rapidly as $T$ increases when $T$ is in $[0.2, 0.3]$, and the success rate of $AE_{sen}$ is always in a high level.
When $T$ reaches $0.3$, the privacy-preserving success rates for $AE_{all}$ and $AE_{sen}$ are 96.1\% and 96.5\%, respectively.
When $T=0.4$, the privacy-preserving success rates for $AE_{all}$ and $AE_{sen}$ are 99.5\% and 99.6\%, respectively.
Generally, for most object detectors, the detection threshold is set to be higher than $0.5$ in order to achieve high detection accuracy.
This indicates that, the proposed method has a good performance in terms of privacy preserving when detection threshold $T$ is set to be greater or equal to $0.3$.

\begin{figure}[!htbp]
	\centering
	\includegraphics[width=2.6in]{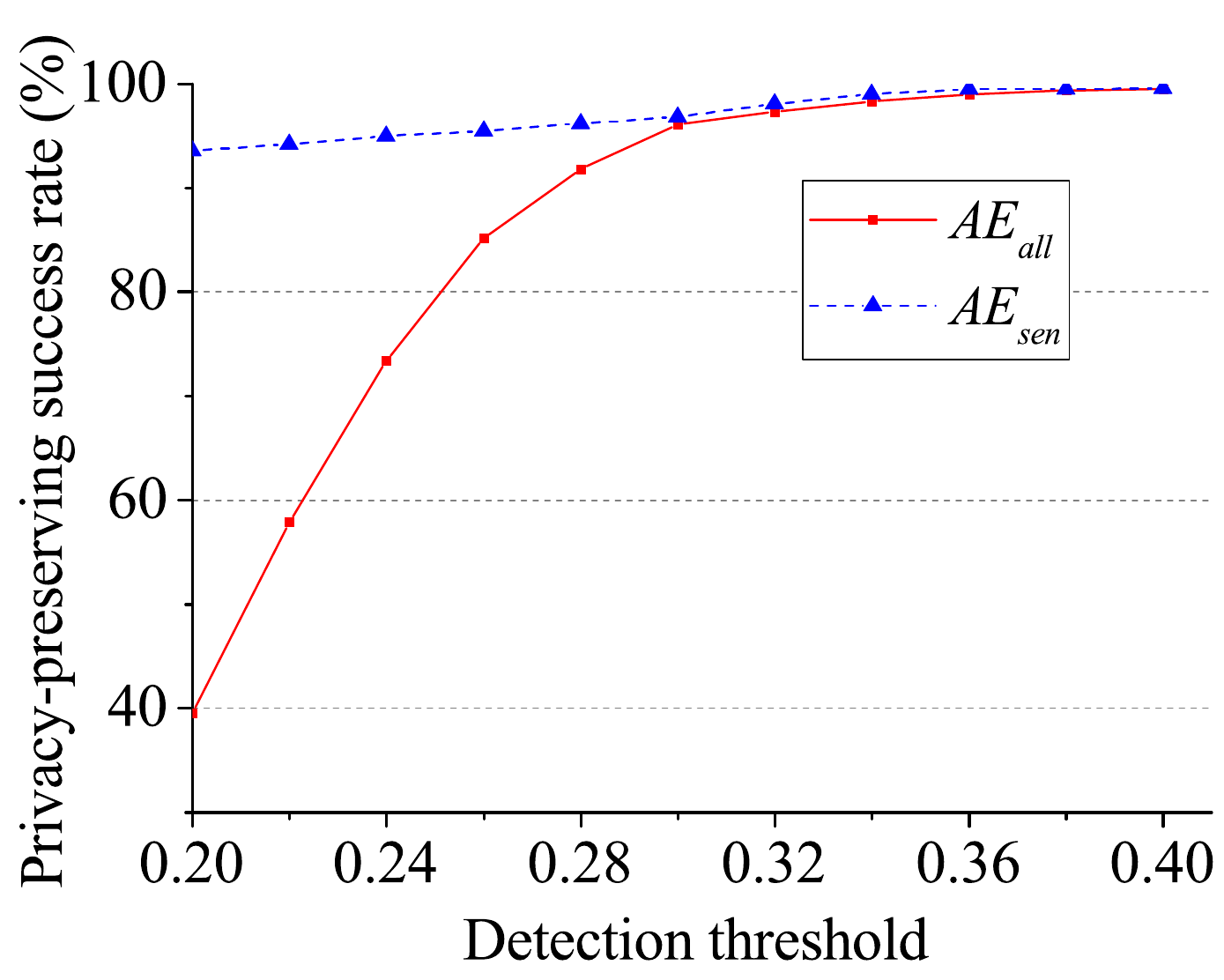}		
	\caption{Privacy-preserving success rate of adversarial examples $AE_{sen}$ under different detection threshold $T$.}
	\label{fig7}
\end{figure}

In addition, we also calculate the privacy leakage rate of the proposed method under different settings of the detection threshold $T$, as shown in TABLE \ref{tab3}.
It is shown that the privacy leakage rate decreases with the increase of $T$.
When $T=0.4$, the privacy leakage rates of $AE_{all}$ and $AE_{sen}$ are 0.06\% and 0.42\%, respectively, which indicate that the proposed method has a good performance in preventing privacy leakage.
Generally, for most object detectors, the detection threshold is set to be higher than 0.5 in order to achieve high detection accuracy.
The result indicates that, the proposed method can effectively reduce the scores of most bounding boxes to below 0.3, which will lead these bounding boxes to be discarded.

\begin{table}[!htbp]
  \centering
  \caption{Privacy leakage rate of adversarial examples under different detection threshold $T$.}
    \begin{tabular}{|c|c|c|c|c|c|c|c|}
    \hline
    \multicolumn{2}{|c|}{Detection threshold} & 0.2 & 0.24 & 0.28 & 0.32 & 0.36 & 0.4 \\
    \hline
    \multirowcell{2}{Privacy\\ leakage rate} & $AE_{all}$ & 8.56\% & 3.84\% & 0.86\% & 0.27\% & 0.12\% & 0.06\% \cr\cline{2-8}
      & $AE_{sen}$ & 4.53\% & 3.36\% & 2.61\% & 1.03\% & 0.50\% & 0.42\% \\
    \hline
    \end{tabular}%
  \label{tab3}%
\end{table}%

\subsection{Comparison with traditional image processing methods} \label{Ex_com}
In this section, we compare the proposed method with the traditional image processing methods.
The traditional image processing methods includes low brightness, noise, mosaic, blur, and JPEG (Joint Photographic Experts Group) compression \cite{wilber2016can, liu2017protecting}.

We select 500 images from MS-COCO dataset to conduct the experiment.
Fig. \ref{fig8} shows an example image processed by different methods and the detection results.
As shown in Fig. \ref{originalimg}, if the image is unprocessed, Faster R-CNN is able to accurately identify all the objects.
As shown in Fig. \ref{lowbrightness}, although low-brightness method seriously degrades the visual quality of the social image, Faster R-CNN can still identify all the objects correctly.
As shown in Fig. \ref{blur}$\sim$Fig. \ref{jpegc}, the persons are all correctly detected in the images processed by blur, mosaic, noise and JPEG compression method, which indicate that these methods are all ineffective to protect the sensitive objects.
As shown in Fig. \ref{ae1} and Fig. \ref{ae2}, Faster R-CNN detects nothing in $AE_{all}$ and recognizes person as dining table in $AE_{sen}$, which indicate that the proposed adversarial example based method can successfully deceive Faster R-CNN.
The reason why the proposed method can resist the detection is that, the feature of a bounding box extracted by Faster R-CNN can be considered as a point in the hyperplane. The proposed method can change the position of the feature point and makes it cross the classification boundary, which will cause the bounding box to be misclassified or be discarded \cite{liu2017protecting}.

\begin{figure*}[h]
\centering
\subfigure[Original image]{
\includegraphics[width=0.46\textwidth]{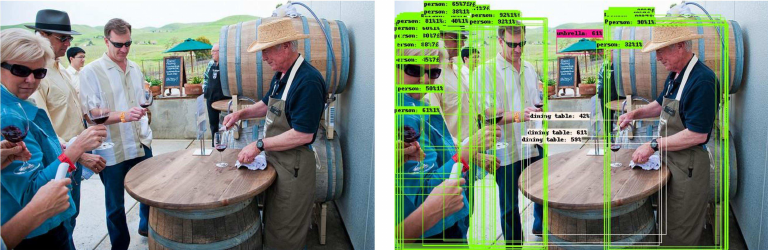}
\label{originalimg}
}
\quad
\subfigure[Low brightness]{
\includegraphics[width=0.46\textwidth]{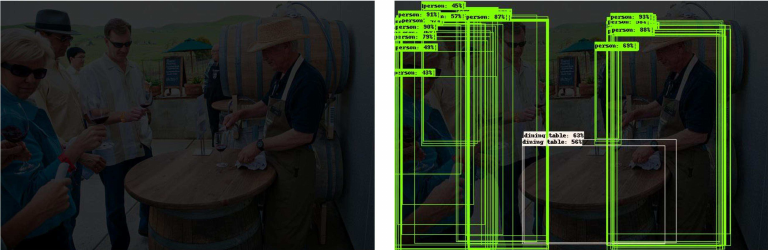}
\label{lowbrightness}
}
\quad
\subfigure[Blur]{
\includegraphics[width=0.46\textwidth]{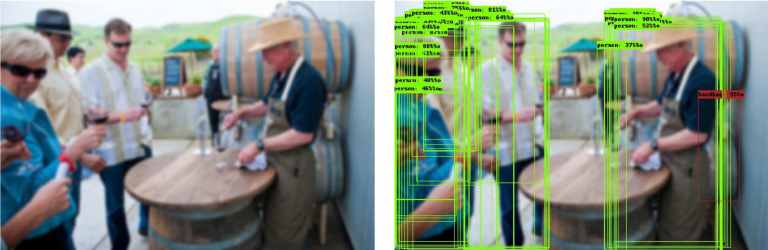}
\label{blur}
}
\quad
\subfigure[Mosaic]{
\includegraphics[width=0.46\textwidth]{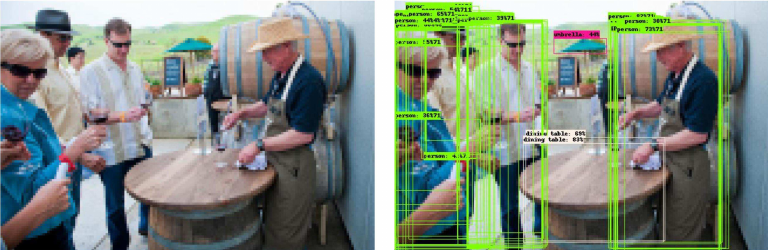}
\label{mosaic}
}
\quad
\subfigure[Noise]{
\includegraphics[width=0.46\textwidth]{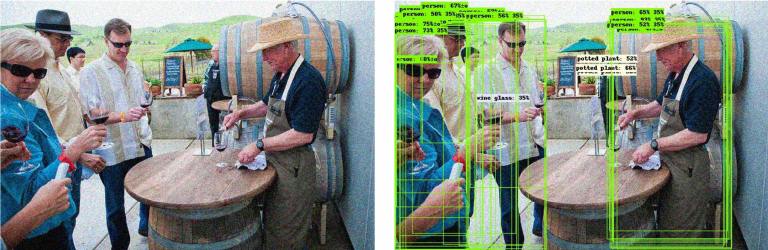}
\label{noise}
}
\quad
\subfigure[JPEG Compression]{
\includegraphics[width=0.46\textwidth]{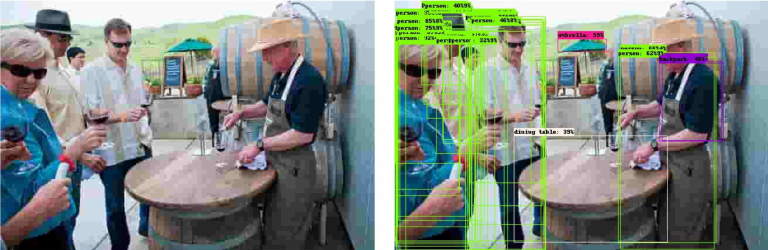}
\label{jpegc}
}
\quad
\subfigure[Adversarial examples $AE_{all}$]{
\includegraphics[width=0.46\textwidth]{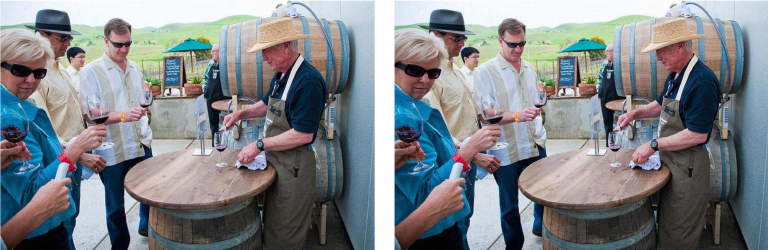}
\label{ae1}
}
\quad
\subfigure[Adversarial examples $AE_{sen}$]{
\includegraphics[width=0.46\textwidth]{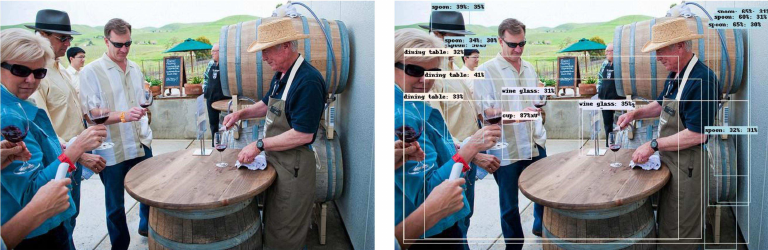}
\label{ae2}
}
\caption{The image example processed by different methods and the detection results from Faster R-CNN.}
\label{fig8}
\end{figure*}

TABLE \ref{tab4} presents the comparison between the proposed method and the above five image processing methods (low brightness, noise, mosaic, blur, and JPEG compression \cite{wilber2016can, liu2017protecting}) in terms of privacy-preserving success rate, privacy leakage rate, PSNR, and SSIM.
The results indicate that the proposed method can effectively protect the privacy of social images as the privacy-preserving success rates of $AE_{all}$ and $AE_{sen}$ are 96.2\% and 96.6\%, respectively, and the privacy leakage rates of $AE_{all}$ and $AE_{sen}$ are 0.54\% and 2.20\%, respectively.
The five traditional image processing methods all fail to protect the social privacy as the privacy-preserving success rates of the five methods are all lower than 5.4\%.
Compared with the five traditional image processing methods, the proposed method has almost no influence on the visual quality of social images as the values of PSNR and SSIM of the proposed method are significantly higher than that of the five traditional image processing methods.
Especially, the values of PSNR and SSIM of the proposed $AE_{sen}$ are 46.91 and 0.992, which are much higher than that of other methods.

\begin{table}[!htbp]
  \centering
  \caption{Comparison between the proposed adversarial example based method and other five image processing methods.}
    \begin{tabular}{ccccc}
    \toprule
    \multicolumn{1}{c}{Methods} & \makecell[c]{Privacy-preserving\\success rate} & \makecell[c]{Privacy\\leakage rate } & \multicolumn{1}{c}{PSNR} & \multicolumn{1}{c}{SSIM} \\
    \midrule
    Low brightness & 0.6\% & 100\% & 8.46  & 0.252 \\
    \midrule
    Blur (Gaussian) & 0.4\% & 100\% & 24.08 & 0.676 \\
    \midrule
    Mosaic & 2.2\% & 78.19\% & 23.87 & 0.683 \\
    \midrule
    Noise  & 0.8\% & 95.55\% & 22.68 & 0.437 \\
    \midrule
    JPEG compression & 5.4\% & 57.32\% & 24.96 & 0.687 \\
    \midrule
    \makecell[c]{The proposed $AE_{all}$} & 96.2\% & 0.54\% & 27.16 & 0.751 \\
    \midrule
    \makecell[c]{The proposed $AE_{sen}$} & 96.6\% & 2.20\% & 46.91 & 0.992 \\
    \bottomrule
    \end{tabular}%
  \label{tab4}%
\end{table}%

\section{Conclusion}  \label{Conclusion}
The malicious attackers can obtain private and sensitive information from uploaded social photos using object detectors.
In this paper, we propose an \textit{Object Disappearance Algorithm} to generate adversarial social images and prevent the object detectors from detecting privacy information.
The proposed method injects imperceptible perturbations to the social images before they are uploaded online.
Specifically, the proposed method generates two kinds of adversarial social images.
One can hide all the objects from being detected by Faster R-CNN, and the other can lead Faster R-CNN to misclassify the customized objects.
We use two metrics, named privacy-preserving success rate and privacy leakage rate, to evaluate the effectiveness of the proposed method, and use PSNR and SSIM to evaluate the impact of the proposed method on the visual quality of social images.
Experimental results show that, compared with traditional image processing methods (low brightness, blur, mosaic, noise, and JPEG compression), the proposed method can effectively resist the detection of Faster R-CNN, while has the minimal influence on the visual quality of social images.
The proposed method iteratively constructs perturbation to generate adversarial social images.
In the future work, we will study the acceleration method for adversarial examples generation to reduce the overhead.

\bibliographystyle{model3-num-names}
\bibliography{ref}

\begin{thebibliography}{34}
\providecommand{\natexlab}[1]{#1}
\providecommand{\url}[1]{\texttt{#1}}
\providecommand{\href}[2]{#2}
\providecommand{\path}[1]{#1}
\providecommand{\eprint}[1]{\href{http://arxiv.org/abs/#1}{\path{#1}}}
\providecommand{\DOIprefix}{doi:}
\providecommand{\ArXivprefix}{arXiv:}
\providecommand{\URLprefix}{URL: }
\providecommand{\Pubmedprefix}{pmid:}
\providecommand{\doi}[1]{\href{http://dx.doi.org/#1}{\path{#1}}}
\providecommand{\Pubmed}[1]{\href{pmid:#1}{\path{#1}}}
\providecommand{\BIBand}{and}
\providecommand{\bibinfo}[2]{#2}
\ifx\xfnm\undefined \def\xfnm[#1]{\unskip,\space#1}\fi
\bibitem[{Hardt and Nath(2012)}]{hardt2012privacy}
\bibinfo{author}{Hardt\xfnm[ M.]}, \bibinfo{author}{Nath\xfnm[ S.]}.
\newblock \bibinfo{title}{Privacy-aware personalization for mobile
  advertising}.
\newblock In: \bibinfo{booktitle}{Proceedings of the {ACM} Conference on
  Computer and Communications Security}, \bibinfo{year}{2012}, pp.
  \bibinfo{pages}{662--673}.
\bibitem[{onl(2017)}]{online2017japan}
\bibinfo{title}{Japan researchers warn of fingerprint theft from `peace' sign}.
\newblock \bibinfo{year}{2017}.
\newblock \URLprefix
  \url{https://phys.org/news/2017-01-japan-fingerprint-theft-peace.html}.
\bibitem[{Klemperer et~al.(2012)Klemperer, Liang, Mazurek, Sleeper, Ur, Bauer
  et~al.}]{klemperer2012tag}
\bibinfo{author}{Klemperer\xfnm[ P.F.]}, \bibinfo{author}{Liang\xfnm[ Y.]},
  \bibinfo{author}{Mazurek\xfnm[ M.L.]}, \bibinfo{author}{Sleeper\xfnm[ M.]},
  \bibinfo{author}{Ur\xfnm[ B.]}, \bibinfo{author}{Bauer\xfnm[ L.]}, et~al.
\newblock \bibinfo{title}{Tag, you can see it! {U}sing tags for access control
  in photo sharing}.
\newblock In: \bibinfo{booktitle}{Proceedings of the {SIGCHI} Conference on
  Human Factors in Computing Systems}, \bibinfo{year}{2012}, pp.
  \bibinfo{pages}{377--386}.
\bibitem[{Vishwamitra et~al.(2017)Vishwamitra, Li, Wang, Hu, Caine and
  Ahn}]{vishwamitra2017towards}
\bibinfo{author}{Vishwamitra\xfnm[ N.]}, \bibinfo{author}{Li\xfnm[ Y.]},
  \bibinfo{author}{Wang\xfnm[ K.]}, \bibinfo{author}{Hu\xfnm[ H.]},
  \bibinfo{author}{Caine\xfnm[ K.]}, \bibinfo{author}{Ahn\xfnm[ G.]}.
\newblock \bibinfo{title}{Towards {PII}-based multiparty access control for
  photo sharing in online social networks}.
\newblock In: \bibinfo{booktitle}{Proceedings of the 22nd {ACM} on Symposium on
  Access Control Models and Technologies}, \bibinfo{year}{2017}, pp.
  \bibinfo{pages}{155--166}.
\bibitem[{Li et~al.(2019{\natexlab{a}})Li, Sun, Li, Niu, Li and
  Cao}]{li2019hideme}
\bibinfo{author}{Li\xfnm[ F.]}, \bibinfo{author}{Sun\xfnm[ Z.]},
  \bibinfo{author}{Li\xfnm[ A.]}, \bibinfo{author}{Niu\xfnm[ B.]},
  \bibinfo{author}{Li\xfnm[ H.]}, \bibinfo{author}{Cao\xfnm[ G.]}.
\newblock \bibinfo{title}{Hideme: {P}rivacy-preserving photo sharing on social
  networks}.
\newblock In: \bibinfo{booktitle}{Proceedings of the {IEEE} Conference on
  Computer Communications}, \bibinfo{year}{2019}{\natexlab{a}}, pp.
  \bibinfo{pages}{154--162}.
\bibitem[{Sun et~al.(2016)Sun, Zhou, Lyu and Zhu}]{sun2016processing}
\bibinfo{author}{Sun\xfnm[ W.]}, \bibinfo{author}{Zhou\xfnm[ J.]},
  \bibinfo{author}{Lyu\xfnm[ R.]}, \bibinfo{author}{Zhu\xfnm[ S.]}.
\newblock \bibinfo{title}{Processing-aware privacy-preserving photo sharing
  over online social networks}.
\newblock In: \bibinfo{booktitle}{Proceedings of the 24th {ACM} Conference on
  Multimedia Conference}, \bibinfo{year}{2016}, pp. \bibinfo{pages}{581--585}.
\bibitem[{Abdulla and Bakiras(2019)}]{abdulla2019hitc}
\bibinfo{author}{Abdulla\xfnm[ A.K.]}, \bibinfo{author}{Bakiras\xfnm[ S.]}.
\newblock \bibinfo{title}{{HITC:} {D}ata privacy in online social networks with
  fine-grained access control}.
\newblock In: \bibinfo{booktitle}{Proceedings of the 24th {ACM} Symposium on
  Access Control Models and Technologies}, \bibinfo{year}{2019}, pp.
  \bibinfo{pages}{123--134}.
\bibitem[{Liu et~al.(2017)Liu, Zhang and Yu}]{liu2017protecting}
\bibinfo{author}{Liu\xfnm[ Y.]}, \bibinfo{author}{Zhang\xfnm[ W.]},
  \bibinfo{author}{Yu\xfnm[ N.]}.
\newblock \bibinfo{title}{Protecting privacy in shared photos via adversarial
  examples based stealth}.
\newblock \bibinfo{journal}{Security and Communication Networks}
  \bibinfo{year}{2017}. \bibinfo{volume}{2017}:\bibinfo{pages}{1--15}.
\bibitem[{Wilber et~al.(2016)Wilber, Shmatikov and Belongie}]{wilber2016can}
\bibinfo{author}{Wilber\xfnm[ M.J.]}, \bibinfo{author}{Shmatikov\xfnm[ V.]},
  \bibinfo{author}{Belongie\xfnm[ S.]}.
\newblock \bibinfo{title}{Can we still avoid automatic face detection}.
\newblock In: \bibinfo{booktitle}{Proceedings of the {IEEE} Winter Conference
  on Applications of Computer Vision}, \bibinfo{year}{2016}, pp.
  \bibinfo{pages}{1--9}.
\bibitem[{Papernot et~al.(2018)Papernot, McDaniel, Sinha and
  Wellman}]{papernot2018sok}
\bibinfo{author}{Papernot\xfnm[ N.]}, \bibinfo{author}{McDaniel\xfnm[ P.]},
  \bibinfo{author}{Sinha\xfnm[ A.]}, \bibinfo{author}{Wellman\xfnm[ M.]}.
\newblock \bibinfo{title}{So{K: S}ecurity and privacy in machine learning}.
\newblock In: \bibinfo{booktitle}{Proceedings of the {IEEE} European Symposium
  on Security and Privacy}, \bibinfo{year}{2018}, pp.
  \bibinfo{pages}{399--414}.
\bibitem[{Akhtar and Mian(2018)}]{akhtar2018threat}
\bibinfo{author}{Akhtar\xfnm[ N.]}, \bibinfo{author}{Mian\xfnm[ A.]}.
\newblock \bibinfo{title}{Threat of adversarial attacks on deep learning in
  computer vision: {A} survey}.
\newblock \bibinfo{journal}{{IEEE} Access} \bibinfo{year}{2018}.
  \bibinfo{volume}{6}:\bibinfo{pages}{14410--14430}.
\bibitem[{Ren et~al.(2017)Ren, He, Girshick and Sun}]{ren2015faster}
\bibinfo{author}{Ren\xfnm[ S.]}, \bibinfo{author}{He\xfnm[ K.]},
  \bibinfo{author}{Girshick\xfnm[ R.B.]}, \bibinfo{author}{Sun\xfnm[ J.]}.
\newblock \bibinfo{title}{Faster {R-CNN}: {T}owards real-time object detection
  with region proposal networks}.
\newblock \bibinfo{journal}{{IEEE} Transactions on Pattern Analysis and Machine
  Intelligence} \bibinfo{year}{2017}.
  \bibinfo{volume}{39}(\bibinfo{number}{6}):\bibinfo{pages}{1137--1149}.
\bibitem[{Lin et~al.(2014)Lin, Maire, Belongie, Hays, Perona, Ramanan
  et~al.}]{lin2014microsoft}
\bibinfo{author}{Lin\xfnm[ T.]}, \bibinfo{author}{Maire\xfnm[ M.]},
  \bibinfo{author}{Belongie\xfnm[ S.J.]}, \bibinfo{author}{Hays\xfnm[ J.]},
  \bibinfo{author}{Perona\xfnm[ P.]}, \bibinfo{author}{Ramanan\xfnm[ D.]},
  et~al.
\newblock \bibinfo{title}{Microsoft {COCO}: {C}ommon objects in context}.
\newblock In: \bibinfo{booktitle}{Proceedings of the European Conference on
  Computer Vision}, \bibinfo{year}{2014}, pp. \bibinfo{pages}{740--755}.
\bibitem[{Everingham et~al.(2010)Everingham, Gool, Williams, Winn and
  Zisserman}]{everingham2010pascal}
\bibinfo{author}{Everingham\xfnm[ M.]}, \bibinfo{author}{Gool\xfnm[ L.V.]},
  \bibinfo{author}{Williams\xfnm[ C.K.I.]}, \bibinfo{author}{Winn\xfnm[ J.M.]},
  \bibinfo{author}{Zisserman\xfnm[ A.]}.
\newblock \bibinfo{title}{The pascal visual object classes ({VOC}) challenge}.
\newblock \bibinfo{journal}{International Journal of Computer Vision}
  \bibinfo{year}{2010}.
  \bibinfo{volume}{88}(\bibinfo{number}{2}):\bibinfo{pages}{303--338}.
\bibitem[{Szegedy et~al.(2014)Szegedy, Zaremba, Sutskever, Bruna, Erhan,
  Goodfellow et~al.}]{szegedy2013intriguing}
\bibinfo{author}{Szegedy\xfnm[ C.]}, \bibinfo{author}{Zaremba\xfnm[ W.]},
  \bibinfo{author}{Sutskever\xfnm[ I.]}, \bibinfo{author}{Bruna\xfnm[ J.]},
  \bibinfo{author}{Erhan\xfnm[ D.]}, \bibinfo{author}{Goodfellow\xfnm[ I.J.]},
  et~al.
\newblock \bibinfo{title}{Intriguing properties of neural networks}.
\newblock In: \bibinfo{booktitle}{Proceedings of the 2nd International
  Conference on Learning Representations}, \bibinfo{year}{2014}, pp.
  \bibinfo{pages}{1--10}.
\bibitem[{Goodfellow et~al.(2015)Goodfellow, Shlens and
  Szegedy}]{goodfellow2014explaining}
\bibinfo{author}{Goodfellow\xfnm[ I.J.]}, \bibinfo{author}{Shlens\xfnm[ J.]},
  \bibinfo{author}{Szegedy\xfnm[ C.]}.
\newblock \bibinfo{title}{Explaining and harnessing adversarial examples}.
\newblock In: \bibinfo{booktitle}{Proceedings of the 3rd International
  Conference on Learning Representations}, \bibinfo{year}{2015}, pp.
  \bibinfo{pages}{1--11}.
\bibitem[{Carlini and Wagner(2017)}]{carlini2017towards}
\bibinfo{author}{Carlini\xfnm[ N.]}, \bibinfo{author}{Wagner\xfnm[ D.]}.
\newblock \bibinfo{title}{Towards evaluating the robustness of neural
  networks}.
\newblock In: \bibinfo{booktitle}{Proceedings of the {IEEE} Symposium on
  Security and Privacy}, \bibinfo{year}{2017}, pp. \bibinfo{pages}{39--57}.
\bibitem[{Chen et~al.(2018)Chen, Cornelius, Martin and
  Chau}]{chen2018shapeshifter}
\bibinfo{author}{Chen\xfnm[ S.]}, \bibinfo{author}{Cornelius\xfnm[ C.]},
  \bibinfo{author}{Martin\xfnm[ J.]}, \bibinfo{author}{Chau\xfnm[ D.H.P.]}.
\newblock \bibinfo{title}{Shapeshifter: {R}obust physical adversarial attack on
  faster {R-CNN} object detector}.
\newblock In: \bibinfo{booktitle}{Proceedings of the Joint European Conference
  on Machine Learning and Knowledge Discovery in Databases},
  \bibinfo{year}{2018}, pp. \bibinfo{pages}{52--68}.
\bibitem[{Brown et~al.(2017)Brown, Man{\'{e}}, Roy, Abadi and
  Gilmer}]{abs-1712-09665}
\bibinfo{author}{Brown\xfnm[ T.B.]}, \bibinfo{author}{Man{\'{e}}\xfnm[ D.]},
  \bibinfo{author}{Roy\xfnm[ A.]}, \bibinfo{author}{Abadi\xfnm[ M.]},
  \bibinfo{author}{Gilmer\xfnm[ J.]}.
\newblock \bibinfo{title}{Adversarial patch}.
\newblock \bibinfo{journal}{arXiv:1712.09665} \bibinfo{year}{2017}.
\bibitem[{Athalye et~al.(2018)Athalye, Engstrom, Ilyas and
  Kwok}]{athalye2017synthesizing}
\bibinfo{author}{Athalye\xfnm[ A.]}, \bibinfo{author}{Engstrom\xfnm[ L.]},
  \bibinfo{author}{Ilyas\xfnm[ A.]}, \bibinfo{author}{Kwok\xfnm[ K.]}.
\newblock \bibinfo{title}{Synthesizing robust adversarial examples}.
\newblock In: \bibinfo{booktitle}{Proceedings of the 35th International
  Conference on Machine Learning}, \bibinfo{year}{2018}, pp.
  \bibinfo{pages}{284--293}.
\bibitem[{Girshick(2015)}]{girshick2015fast}
\bibinfo{author}{Girshick\xfnm[ R.]}.
\newblock \bibinfo{title}{Fast {R-CNN}}.
\newblock In: \bibinfo{booktitle}{Proceedings of the {IEEE} International
  Conference on Computer Vision}, \bibinfo{year}{2015}, pp.
  \bibinfo{pages}{1440--1448}.
\bibitem[{Xue et~al.(2019)Xue, Yuan, Sun and Wu}]{ourpatent}
\bibinfo{author}{Xue\xfnm[ M.]}, \bibinfo{author}{Yuan\xfnm[ C.]},
  \bibinfo{author}{Sun\xfnm[ S.]}, \bibinfo{author}{Wu\xfnm[ Z.]}.
\newblock \bibinfo{title}{A privacy protection method and device for social
  photos against yolo object detector}.
\newblock \bibinfo{year}{2019}.
\newblock \bibinfo{note}{Chinese Patent, Patent No. 201911346202.0, Filed
  December, 2019}.
\bibitem[{Li et~al.(2019{\natexlab{b}})Li, Shamsabadi, Sanchez{-}Matilla,
  Mazzon and Cavallaro}]{li2019scene}
\bibinfo{author}{Li\xfnm[ C.Y.]}, \bibinfo{author}{Shamsabadi\xfnm[ A.S.]},
  \bibinfo{author}{Sanchez{-}Matilla\xfnm[ R.]}, \bibinfo{author}{Mazzon\xfnm[
  R.]}, \bibinfo{author}{Cavallaro\xfnm[ A.]}.
\newblock \bibinfo{title}{Scene privacy protection}.
\newblock In: \bibinfo{booktitle}{Proceedings of the {IEEE} International
  Conference on Acoustics, Speech and Signal Processing},
  \bibinfo{year}{2019}{\natexlab{b}}, pp. \bibinfo{pages}{2502--2506}.
\bibitem[{Shen et~al.(2019)Shen, Fan, Wong, Ng and Kankanhalli}]{ShenFWNK19}
\bibinfo{author}{Shen\xfnm[ Z.]}, \bibinfo{author}{Fan\xfnm[ S.]},
  \bibinfo{author}{Wong\xfnm[ Y.]}, \bibinfo{author}{Ng\xfnm[ T.]},
  \bibinfo{author}{Kankanhalli\xfnm[ M.S.]}.
\newblock \bibinfo{title}{Human-imperceptible privacy protection against
  machines}.
\newblock In: \bibinfo{booktitle}{Proceedings of the 27th {ACM} International
  Conference on Multimedia}, \bibinfo{year}{2019}, pp.
  \bibinfo{pages}{1119--1128}.
\bibitem[{Shan et~al.(2020)Shan, Wenger, Zhang, Li, Zheng and
  Zhao}]{abs-2002-08327}
\bibinfo{author}{Shan\xfnm[ S.]}, \bibinfo{author}{Wenger\xfnm[ E.]},
  \bibinfo{author}{Zhang\xfnm[ J.]}, \bibinfo{author}{Li\xfnm[ H.]},
  \bibinfo{author}{Zheng\xfnm[ H.]}, \bibinfo{author}{Zhao\xfnm[ B.Y.]}.
\newblock \bibinfo{title}{Fawkes: {P}rotecting personal privacy against
  unauthorized deep learning models}.
\newblock \bibinfo{journal}{arXiv:2002.08327} \bibinfo{year}{2020}.
\bibitem[{Zhao et~al.(2018)Zhao, Zhu, Liang, Shen, Zhang and
  Chen}]{zhao2018seeing}
\bibinfo{author}{Zhao\xfnm[ Y.]}, \bibinfo{author}{Zhu\xfnm[ H.]},
  \bibinfo{author}{Liang\xfnm[ R.]}, \bibinfo{author}{Shen\xfnm[ Q.]},
  \bibinfo{author}{Zhang\xfnm[ S.]}, \bibinfo{author}{Chen\xfnm[ K.]}.
\newblock \bibinfo{title}{Seeing isn't believing: {P}ractical adversarial
  attack against object detectors}.
\newblock \bibinfo{journal}{arXiv:1812.10217} \bibinfo{year}{2018}.
\bibitem[{Song et~al.(2018)Song, Eykholt, Evtimov, Fernandes, Li, Rahmati
  et~al.}]{eykholt2018physical}
\bibinfo{author}{Song\xfnm[ D.]}, \bibinfo{author}{Eykholt\xfnm[ K.]},
  \bibinfo{author}{Evtimov\xfnm[ I.]}, \bibinfo{author}{Fernandes\xfnm[ E.]},
  \bibinfo{author}{Li\xfnm[ B.]}, \bibinfo{author}{Rahmati\xfnm[ A.]}, et~al.
\newblock \bibinfo{title}{Physical adversarial examples for object detectors}.
\newblock In: \bibinfo{booktitle}{Proceedings of the 12th {USENIX} Workshop on
  Offensive Technologies}, \bibinfo{year}{2018}, pp. \bibinfo{pages}{1--10}.
\bibitem[{Redmon et~al.(2016)Redmon, Divvala, Girshick and
  Farhadi}]{redmon2016you}
\bibinfo{author}{Redmon\xfnm[ J.]}, \bibinfo{author}{Divvala\xfnm[ S.K.]},
  \bibinfo{author}{Girshick\xfnm[ R.]}, \bibinfo{author}{Farhadi\xfnm[ A.]}.
\newblock \bibinfo{title}{You only look once: {U}nified, real-time object
  detection}.
\newblock In: \bibinfo{booktitle}{Proceedings of the {IEEE} Conference on
  Computer Vision and Pattern Recognition}, \bibinfo{year}{2016}, pp.
  \bibinfo{pages}{779--788}.
\bibitem[{Liu et~al.(2016)Liu, Anguelov, Erhan, Szegedy, Reed, Fu
  et~al.}]{liu2016ssd}
\bibinfo{author}{Liu\xfnm[ W.]}, \bibinfo{author}{Anguelov\xfnm[ D.]},
  \bibinfo{author}{Erhan\xfnm[ D.]}, \bibinfo{author}{Szegedy\xfnm[ C.]},
  \bibinfo{author}{Reed\xfnm[ S.E.]}, \bibinfo{author}{Fu\xfnm[ C.]}, et~al.
\newblock \bibinfo{title}{{SSD}: {S}ingle shot multibox detector}.
\newblock In: \bibinfo{booktitle}{Proceedings of the European Conference on
  Computer Vision}, \bibinfo{year}{2016}, pp. \bibinfo{pages}{21--37}.
\bibitem[{Dai et~al.(2016)Dai, Li, He and Sun}]{dai2016r}
\bibinfo{author}{Dai\xfnm[ J.]}, \bibinfo{author}{Li\xfnm[ Y.]},
  \bibinfo{author}{He\xfnm[ K.]}, \bibinfo{author}{Sun\xfnm[ J.]}.
\newblock \bibinfo{title}{{R-FCN}: {O}bject detection via region-based fully
  convolutional networks}.
\newblock In: \bibinfo{booktitle}{Proceedings of the 30th International
  Conference on Neural Information Processing Systems}, \bibinfo{year}{2016},
  pp. \bibinfo{pages}{379--387}.
\bibitem[{Huang et~al.(2017)Huang, Rathod, Sun, Zhu, Korattikara, Fathi
  et~al.}]{huang2017speed}
\bibinfo{author}{Huang\xfnm[ J.]}, \bibinfo{author}{Rathod\xfnm[ V.]},
  \bibinfo{author}{Sun\xfnm[ C.]}, \bibinfo{author}{Zhu\xfnm[ M.]},
  \bibinfo{author}{Korattikara\xfnm[ A.]}, \bibinfo{author}{Fathi\xfnm[ A.]},
  et~al.
\newblock \bibinfo{title}{Speed/accuracy trade-offs for modern convolutional
  object detectors}.
\newblock In: \bibinfo{booktitle}{Proceedings of the {IEEE} Conference on
  Computer Vision and Pattern Recognition}, \bibinfo{year}{2017}, pp.
  \bibinfo{pages}{3296--3297}.
\bibitem[{Szegedy et~al.(2016)Szegedy, Vanhoucke, Ioffe, Shlens and
  Wojna}]{szegedy2016rethinking}
\bibinfo{author}{Szegedy\xfnm[ C.]}, \bibinfo{author}{Vanhoucke\xfnm[ V.]},
  \bibinfo{author}{Ioffe\xfnm[ S.]}, \bibinfo{author}{Shlens\xfnm[ J.]},
  \bibinfo{author}{Wojna\xfnm[ Z.]}.
\newblock \bibinfo{title}{Rethinking the inception architecture for computer
  vision}.
\newblock In: \bibinfo{booktitle}{Proceedings of the {IEEE} Conference on
  Computer Vision and Pattern Recognition}, \bibinfo{year}{2016}, pp.
  \bibinfo{pages}{2818--2826}.
\bibitem[{Abadi et~al.(2016)Abadi, Agarwal, Barham, Brevdo, Chen, Citro
  et~al.}]{abadi2016tensorflow}
\bibinfo{author}{Abadi\xfnm[ M.]}, \bibinfo{author}{Agarwal\xfnm[ A.]},
  \bibinfo{author}{Barham\xfnm[ P.]}, \bibinfo{author}{Brevdo\xfnm[ E.]},
  \bibinfo{author}{Chen\xfnm[ Z.]}, \bibinfo{author}{Citro\xfnm[ C.]}, et~al.
\newblock \bibinfo{title}{Tensorflow: {L}arge-scale machine learning on
  heterogeneous distributed systems}.
\newblock \bibinfo{journal}{arXiv:1603.04467} \bibinfo{year}{2016}.
\bibitem[{Wang et~al.(2004)Wang, Bovik, Sheikh and Simoncelli}]{WangBSS04}
\bibinfo{author}{Wang\xfnm[ Z.]}, \bibinfo{author}{Bovik\xfnm[ A.C.]},
  \bibinfo{author}{Sheikh\xfnm[ H.R.]}, \bibinfo{author}{Simoncelli\xfnm[
  E.P.]}.
\newblock \bibinfo{title}{Image quality assessment: {F}rom error visibility to
  structural similarity}.
\newblock \bibinfo{journal}{{IEEE} Transactions on Image Processing}
  \bibinfo{year}{2004}.
  \bibinfo{volume}{13}(\bibinfo{number}{4}):\bibinfo{pages}{600--612}.

\end{thebibliography}

\end{document}